# Machine Learning Risk Intelligence for Green Hydrogen Investment: Insights for Duqm R3 Auction

Obumneme Nwafor[1] and Mohammed Abdul Majeed Al Hooti[2]

[1]School of Computing, Engineering and Built Environment, Glasgow Caledonian University, Scotland
ORCID: 0000-0002-0993-1659.

[2] Department of Business Management, University of Lancashire United Kingdom
ORCID: 0009-0005-5793-7122

**Abstract**
As green hydrogen emerges as a major component of global decarbonisation, Oman has positioned itself strategically through national auctions and international partnerships. Following two successful green hydrogen project rounds, the country launched its third auction (R3) in the Duqm region. While this area exhibits relative geospatial homogeneity, it is still vulnerable to environmental fluctuations that pose inherent risks to productivity. Despite growing global investment in green hydrogen, operational data remains scarce, with major projects like Saudi Arabia's NEOM facility not expected to commence production until 2026, and Oman's ACME Duqm project scheduled for 2028. This absence of historical maintenance and performance data from large-scale hydrogen facilities in desert environments creates a major knowledge gap for accurate risk assessment for infrastructure planning and auction decisions. Given this data void, environmental conditions emerge as accessible and reliable proxy for predicting infrastructure maintenance pressures, because harsh desert conditions such as dust storms, extreme temperatures, and humidity fluctuations are well-documented drivers of equipment degradation in renewable energy systems. To address this challenge, this paper proposes an Artificial Intelligence decision support system that leverages publicly available meteorological data to develop a predictive Maintenance Pressure Index (MPI), which predicts risk levels and future maintenance demands on hydrogen infrastructure. This tool strengthens regulatory foresight and operational decision-making by enabling temporal benchmarking to assess and validate performance claims over time. It can be used to incorporate temporal risk intelligence into auction evaluation criteria despite the absence of historical operational benchmarks.

**Corresponding author**: Obumneme Nwafor obumnemenwafor@gmail.com

# 1. Introduction

As countries accelerate their shift to clean energy, green hydrogen has emerged as a major solution for decarbonisation and attainment of net-zero goals. Produced via water electrolysis powered by renewables, it enables clean energy to be stored, transported, and used as industrial feedstock, replacing processes that traditionally rely on fossil fuels. This transformation has resulted in a surge in global investment, with the International Energy Agency forecasting a $1.2 trillion hydrogen market by 2030[1]. Oman has emerged as a strategic player in the global hydrogen economy, leveraging its abundant solar and wind resources, strategic geographic location, and established energy infrastructure. The Sultanate's hydrogen strategy encompasses three successive auction rounds, with the third round (R3) offering approximately 300 km² of land in the Duqm region for green hydrogen development [2]. This initiative represents one of the largest single land allocations for hydrogen projects globally, signalling Oman's commitment to becoming a major hydrogen exporter. However, the success of large-scale hydrogen infrastructure investments not only depends on operational reliability, but also on maintenance efficiency.

Environmental factors, particularly in arid regions like Duqm, can significantly impact equipment performance, maintenance schedules, and overall project economics. Sand storms, temperature extremes, humidity variations, and wind patterns directly influence the operational efficiency of electrolysers, solar panels, wind turbines, and associated infrastructure components [3]. However, the nascent state of commercial green hydrogen production presents a fundamental data scarcity challenge. For instance, flagship projects such as NEOM's 600 tonnes/day facility in Saudi Arabia [4] and Oman's 497 ktpa ACME Duqm project [5] represent the first wave of mega-scale developments, but neither of the two projects is expected to start operations and generate operational data until 2026 and 2028 respectively. This temporal gap between current investment decisions and future operational validation necessitates the development of predictive risk assessment frameworks that can operate effectively in data-sparse environments and account for environmental stressors.

This paper addresses the gap in environmental risk assessment for hydrogen infrastructure by developing a machine learning-based Maintenance Pressure Index (MPI) specifically designed for Oman's Duqm R3 auction zone. Current investment evaluation frameworks for hydrogen projects typically adopt deterministic approaches that may not fully accommodate the evolving complexities of environmental risks [6]. This paper addresses these challenges by introducing AI-based risk intelligence framework specifically designed for green hydrogen investment evaluation. The core contribution is the development of a composite Maintenance Pressure Index (MPI) that quantifies weather-induced maintenance pressures on hydrogen infrastructure, providing stakeholders with insights for climate-related time-sensitive risk management and operational planning. It provides risk assessment methods that offer temporal granularity and stronger predictive capabilities to support infrastructure planning and regulatory decision-making. To contextualize this research within existing knowledge, we review three key areas: hydrogen auction mechanisms, environmental risk factors, and AI applications in infrastructure planning.

## 2. Literature Review - Hydrogen Auctions, Environmental Risk and AI

As global interest in green hydrogen intensifies, governments are increasingly turning to competitive auction mechanisms as a means of allocating land, resources, and incentives for hydrogen production. These auctions serve not only to attract private investment but also to promote price discovery, and technology diversification through transparent and market-driven selection processes [7]. Similar to the well-established practice in renewable energy procurement, hydrogen auctions represent a major policy tool for stimulating early market development and scaling infrastructure investment.

In emerging hydrogen economies such as Oman, the auction model plays a foundational role in structuring the transition from pilot projects to utility-scale hydrogen production. The government, through Hydrogen Oman (Hydrom), has launched successive auction rounds offering exclusive development zones with co-located access to solar and wind resources, land parcels, and desalinated water supply [2]. These auctions establish the minimum technical criteria and financial commitments to ensure project viability and alignment with national energy transition objectives. However, unlike mature electricity auctions, the evolving nature of hydrogen value chains introduces multidimensional investment risks that are often underrepresented in current auction frameworks [8].

Investment risks in hydrogen infrastructure development can be broadly categorized into four interrelated domains: technical, financial, regulatory, and environmental [9] [10][11]. Technical risks relate to the performance, scalability, and integration of electrolyser systems, particularly under harsh climatic conditions [12]. Financial risks refer to capital intensity, revenue uncertainties due to off-take arrangements, and evolving market prices for hydrogen [13]. Regulatory risks include licencing delays, inconsistent policy incentives, and long-term uncertainty in emission reduction credit frameworks. Environmental risk deals with climatic stressors such as high temperatures, dust storms, wind variability, and extreme irradiance which can substantially increase system downtime and compromise the durability of power electronics and desalination units [14][15]. They are often the least quantified yet most operationally impactful. Research by [16] [17] and [18] demonstrated that temperatures exceeding 45°C impose thermal stress on, and can reduce electrolyser efficiency by up to 15%. Similarly, dust accumulation on solar panels can decrease electricity generation by 20-40% in arid regions [19]. Although wind is considered a complementary renewable energy source in hydrogen production, its variability introduces mechanical and structural challenges. Sudden gusts and high turbulence intensities can cause blade misalignment, yaw system failures, and accelerated bearing wear in wind turbines. Furthermore, wind-blown sand particles contribute to abrasion of exposed surfaces, such as PV panels, pipelines, and condenser coils.

The Middle East's harsh environmental conditions present unique challenges for hydrogen infrastructure. [20] identified sand storms as a primary cause of equipment degradation in Saudi Arabia's NEOM hydrogen project, leading to increased maintenance costs and reduced operational availability. Similarly, studies on Qatar's hydrogen initiatives revealed that humidity fluctuations significantly impact the performance of power electronics and control systems[20]. Globally, there is growing recognition of the need to evolve hydrogen auction designs to account for location-specific risks and metrics. For example, Chile and Australia have begun incorporating environmental impact assessments and risk-readiness indicators into their hydrogen project evaluations [21] [22]. However, a standardized methodology for integrating dynamic environmental risk intelligence into hydrogen auction planning is still lacking.

The increasing complexity of infrastructure systems and the growing availability of high-resolution environmental data have made AI a transformative tool in predictive risk modelling. In the context of green hydrogen and spatially extensive projects like Duqm R3, environmental risk must be analysed across both space and time. AI models such as spatiotemporal neural networks and ensemble learning methods, have proven effective in capturing complex environmental dynamics [22] [23]. These models enable the forecasting of future risk exposures and can simulate operational conditions under various climate scenarios. Time-series prediction algorithms such as Long Short-Term Memory (LSTM) networks and Temporal Convolutional Networks (TCNs) are increasingly applied for their ability to retain temporal dependencies and model nonlinear climate trends [24]. When combined with satellite-derived data, these models can provide granular insight into the environmental risk landscape.

To address these challenges, this study proposes that hydrogen auctions, particularly in climatically volatile regions such as Duqm, should include AI-based environmental risk models as part of the auction or evaluation process. Omission of such risks models from auction scoring criteria may result in the selection of technically viable but operationally vulnerable projects. Such models can quantify the latent maintenance stress that infrastructure may face over its lifecycle. From an investor's perspective, the absence of transparent, forward-looking environmental risk indicators can undermine the accuracy of financial models, affect insurance premiums, and increase the cost of capital. Banks and multilateral financiers, who typically require environmental risk disclosures under frameworks such as the Equator Principles and Task Force on Climate-related Financial Disclosures (TCFD), may view this gap as a material barrier to investment [25]. Building on the identified research gaps, this study develops a comprehensive AI framework that integrates environmental risk assessment into hydrogen infrastructure planning.

## 3. Methodology

This study adopts a hybrid predictive AI pipeline consisting of classification, explainability, and time series forecasting. The goal is to produce an MPI that supports strategic infrastructure investment and risk-informed auction design in the Duqm R3 hydrogen development zone. The pipeline starts with an XGBoost model trained to classify maintenance risk levels based on environmental and meteorological drivers. SHAP is applied to the XGBoost model to reveal the relative influence of each factor. Finally, the most influential variables are passed into a Prophet model, which captures long-term trends and seasonality to forecast weekly maintenance scores. The following sections presents the study area, data processing workflow, feature engineering, model architecture, and performance evaluation approach. The schematic diagram of this study methodology is given in Figure 1 below

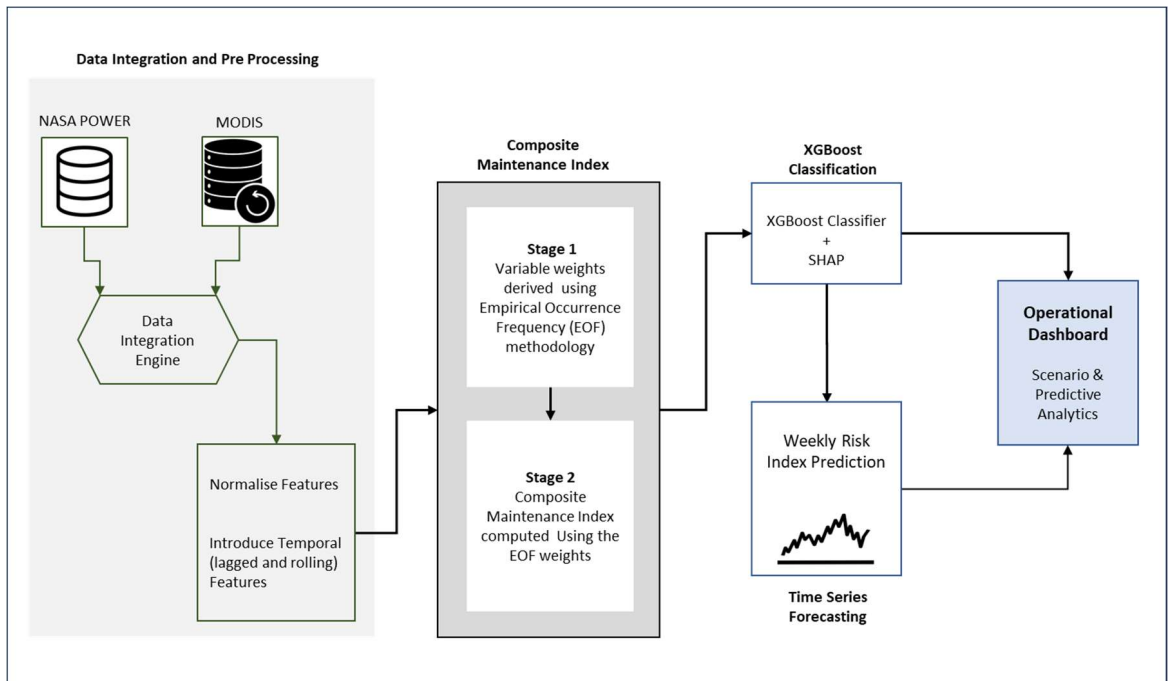

**Figure 1** Schematic diagram of study methodology

### 3.1 Study Area: Duqm R3 Hydrogen Auction Zone

The Duqm Special Economic Zone (SEZ), located on Oman's south-eastern coast along the Arabian Sea, has been selected as the study area due to its designation as a green hydrogen development hub in the national hydrogen auction framework. While the region benefits from strong solar irradiance and consistent wind profiles, it is also subject to extreme environmental conditions, including frequent dust storms, high temperatures exceeding 45°C in summer, and wind-driven particulates. These conditions pose substantial risks to hydrogen infrastructure in the absence of real-time predictive maintenance intelligence.

### 3.2 Data Sources

The environmental variables used in this study were selected based on their relevance to infrastructure maintenance risk. Data were sourced from publicly available, high-resolution earth observation platforms. First, daily meteorological data such as including solar irradiance, temperature, and wind speed was obtained from NASA's Prediction of Worldwide Energy Resources (POWER) project. This is followed by capturing the Aerosol Optical Depth (AOD) from the Moderate Resolution Imaging Spectroradiometer (MODIS) and European Space Agency's Sentinel-5P Tropospheric Monitoring Instrument (TROPOMI) collections, at 0.47 μm spectral band and spatial resolution of 1 km²; buffer radius centred on the coordinates of Duqm. The resulting dataset provides a representation of Duqm's atmospheric conditions relevant to green hydrogen production, from January 2020 to December 2024. Table 1 below shows a list and description of the variables in the dataset.

**Table 1.** Dataset Feature Names, Units, Description and Relevance

| Feature Name | Unit / Data Type | Description | Relevance(Why it matters) |
|---|---|---|---|
| Solar Irradiance | W/m² (Watts per square meter) | Measures the power of solar radiation received per unit area. | Critical for solar-powered energy in green hydrogen systems; high irradiance improves solar PV/electrolysis efficiency. |
| Temperature | °C (Degrees Celsius) | Daily average or maximum air temperature at 2 meters above ground. | Influences electrolysis efficiency, water evaporation rates, and cooling requirements for equipment |
| Wind Speed | m/s (Meters per second) | Near-surface wind speed, typically at 10 meters elevation. | Enables integration of wind energy into hybrid renewable hydrogen systems; affects turbine output and feasibility of off-grid operations. |
| AOD (Aerosol Optical Depth) | Unitless (dimensionless) | Represents the degree to which aerosols prevent light transmission; values typically range from 0 to >3. | High AOD reduces solar irradiance, affecting hydrogen production yield and system efficiency; also indicates potential maintenance due to dust accumulation. |
| Relative Humidity | % | A measure of the atmospheric moisture content | Excessive moisture is linked to corrosion risk in exposed metallic components and insulation degradation |
| Month | Unitless (dimensionless) | Month of the year | Captures seasonal variability in irradiance, temperature, and AOD; enables modelling of temporal effects on hydrogen yield and resource planning. |

### 3.3 Maintenance Pressure Index, Label and Thresholds

Due to limited historical maintenance logs or infrastructure failure datasets for green hydrogen production systems in Duqm, a heuristic approach was used to derive the MPI based on literature-backed performance thresholds. AOD levels above 0.9 are associated with particulate matter and dust storms, which increase PV soiling and filtration load, leading to pressure losses and maintenance needs [26][27]. Temperature above 25°C causes measurable efficiency loss through increased carrier recombination rates. For every 1°C increase above 25°C, solar panels lose 0.3% to 0.5% of their efficiency [28][29]. Also, relative humidity above 80% is linked to corrosion risk in exposed metallic components and insulation degradation [30] [31]. Similarly, while catastrophic failure occurs at very high wind speeds, subtle performance impacts from vibration and dust accumulation can begin at moderate wind speeds such as 5m/s, but the exact threshold depends on local environmental conditions, panel mounting systems, and dust particle characteristics [26] [32]. Also, high fluctuation in irradiance impacts power stability for electrolyser stacks [33]. In this study, each of the above conditions was modelled as a binary risk trigger (1 if threshold exceeded, else 0), and a composite score was calculated using Equation 1 below.

$$MPI = \sum_{i=1}^{n} w_i \cdot f_i(x_i) \quad (1)$$

Where $x_i$ is the *i-th environmental factor* (e.g., AOD, temperature, humidity), $w_i$ is the *weight* assigned $x_i$, and $f_i(x_i)$ is the transformation function that maps the input variable to a maintenance risk score (binary via a threshold). Example of the transformation is given in equation (2) below

$$f_i(x_i) = \begin{cases} 1, & \text{if } x_i > t \\ 0, & \text{otherwise} \end{cases} \quad (2)$$

Using the variables in Table 1 above, the MPI is given by equation (3) below

$$MPI = w_1 \cdot w_2 \cdot \|(Temp > \tau_2) + w_3 \cdot \|(Humidity > \tau_3) + w_4 \cdot \|(Wind > \tau_4) + w_5 \cdot \|(IrrVar > \tau_5) \quad (3)$$

Where $\tau_i$ is the thresholds for each variable, IrrVar is the 3-day rolling standard deviation of irradiance and $\sum w_i = 1$. The weights ($w_i$) were determined using the Empirical Occurrence Frequency (EOF) methodology which computes the rate at which each risk condition (i.e. the variable is above the threshold) was observed in the Duqm dataset. Table 2 below show the parameters used for the CMI.

Table 2. EOF weights and risk threshold for environmental variables

| Variable | EOF weight | Risk Threshold |
|---|---|---|
| AOD | 0.35 | 0.9 |
| Temperature | 0.25 | 35 |
| Humidity | 0.20 | 70 |
| Wind Speed | 0.15 | 5 |
| Solar Irradiance 3d STD | 0.05 | 90th percentile |

### 3.4 Maintenance Pressure Labelling
To enable supervised learning for risk prediction, we converted the continuous MPI score into Risk Categories based on 25, 50 and 75 percentiles as shown in table 3 below:

Table 3 Risk Categories based on CMI Threshold

| Risk Category | MPI | Description |
|---|---|---|
| High | >=0.6 | High-likelihood maintenance alert zone. |
| Medium | 0.3=< score <0.6 | Moderate risk accumulation that may require condition-based monitoring. |
| Low | < 0.3 | Favourable operational conditions with minimal intervention needs. |

### 3.5 Feature Engineering and Variable Transformation
The dataset was normalised using the MinMax Scaler represented by Equation (4) below:

$$\hat{X}_{ij} = \frac{X_{ij} - \min(X_{.j})}{\max(X_{.j}) - \min(X_{.j})} \quad (4)$$

Where $X_{ij}$ is the original value of feature j for day I, $\min(X_{.j})$ is the minimum value of feature j across the entire training set, $\max(X_{.j})$ represents the maximum value of feature j across the entire training set and $\hat{X}_{ij}$ is the scaled value constrained to the interval [0,1]. Directional transformations were also applied where appropriate (e.g., 1 - AOD) to reflect features with maintenance impact logic. Additionally, rolling statistical features were engineered to capture the following:
   a. Immediate Impact Variables : aod, solar_irradiance, temperature, humidity, wind_speed
   b. Cumulative Impact Variables: aod_rolling_3d_mean, aod_rolling_7d_mean, irradiance_rolling_3d_mean, irradiance_rolling_7d_mean
   c. Variability Features: aod_rolling_3d_std, aod_rolling_7d_std, irradiance_rolling_3d_std, irradiance_rolling_7d_std

### 3.6 XGBoost Modelling for Surrogate Explainability and Feature Profiling

As a precursor to the time-series forecasting, XGBoost model was developed to serve as a surrogate model for capturing non-linear relationships and identifying influential predictors of weekly maintenance scores. The model was trained on engineered features derived from historical environmental data, including lagged, rolling, and smoothed versions of AOD, temperature, humidity, and wind speed. Feature values were standardized using MinMax scaling to ensure numerical stability and improve convergence. Subsequently, SHAP were applied to the XGBoost model to quantify the contribution of each input feature to the model's output. For the next stage of model pipeline, the most predictive features, as revealed by SHAP, were used as input regressors for the Prophet forecasting model to capture long-term trend and seasonal patterns in maintenance pressure.

### 3.7 Time Series Forecasting Using Prophet Model

To forecast the future trajectory of weekly maintenance scores, this study employed the Facebook Prophet model [34], a generalized additive model (GAM) framework designed for interpretable time-series forecasting. Prophet is particularly suited for time-series data exhibiting multiple seasonality, abrupt change points, and holidays or external regressors. Prophet models a time-series $y(t)$ as an additive combination of trend, seasonality, holiday effects, and exogenous regressors pressed by equation 5 below:

$$y(t) = g(t) + s(t) + h(t) + \varepsilon_t \quad (5)$$

Where $g(t)$ is the piecewise linear or logistic growth trend, $s(t)$ is the periodic seasonality (e.g., yearly), $h(t)$ is the effects of holidays or known events and $\varepsilon_t$ represents error term assumed to be normally distributed. Additional regressors $x(t)$ can be included as

$$y(t) = g(t) + s(t) + h(t) + \beta x(t) + \varepsilon_t \quad (6)$$

In this study, the regressors are the lagged and rolling features of AOD, Temperature, Humidity and Wind Speed. The model was trained on weekly mean maintenance scores, resampled from daily data to reduce noise and align with maintenance planning cycles.

## 3.8 Forecast Horizon and Evaluation

The model was extended into the future for a user-specified forecast horizon $H \in \{4,12,52\}$. The latest available exogenous feature values were forward-filled to generate the future design matrix required for forecasting.

## 4. Results and Discussion

### 4.1 Exploratory Data Analysis

Extensive exploratory data analysis (EDA) was conducted to provide insights into the distributions, relationships, and temporal patterns among the variables. First, to understand seasonal patterns and trends in dust exposure, the AOD time series was decomposed into trend, seasonality, and residual components using a 30-day periodicity as shown in Figure 2 below. This decomposition was useful in identifying long-term dust build-up patterns that contribute to maintenance risk.

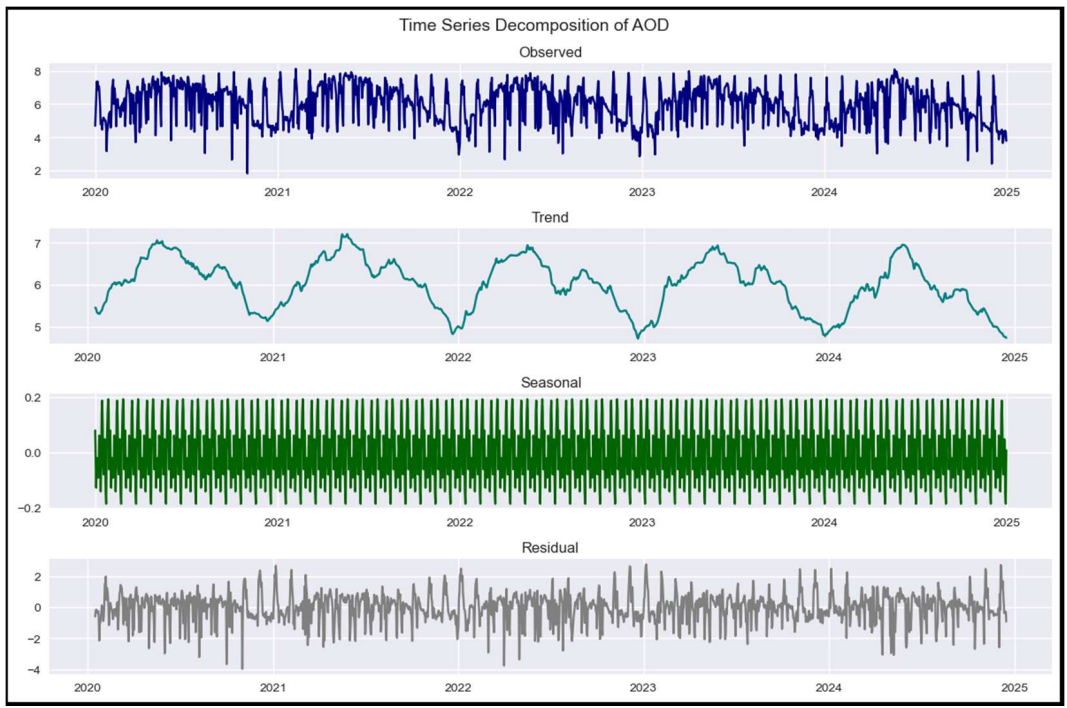

**Figure 2** Time Series Decomposition of AOD

Figure 2 show AOD values with noticeable spikes and dips throughout the time series. These fluctuations suggest episodic events such as dust storms or clear-sky intervals indicating the efficiency of hydrogen production is expected to vary. The trend line shows distinct seasonal cycles with observable peaks occurring approximately once a year. This indicates a recurring build-up and dissipation of aerosol load, possibly linked to seasonal climatic conditions like monsoon onset, dry seasons, or wind patterns. The seasonal pattern is highly periodic and regular, suggesting strong annual or sub-annual cycles in aerosol behaviour. The residual captures short-term anomalies and noise not explained by trend or seasonality. It shows that there are several strong outliers (both positive and negative), which suggest that there are extreme AOD deviations that may correspond to abrupt dust storm events or unusual atmospheric clearing. These anomalies imply that there are periods when infrastructure maintenance rates deviate from baseline expectations, hence the need for a robust adaptive risk management plan.

## 4.2 XGBoost Model Evaluation

Figure 3 shows the confusion matrix of the three-class classification model.

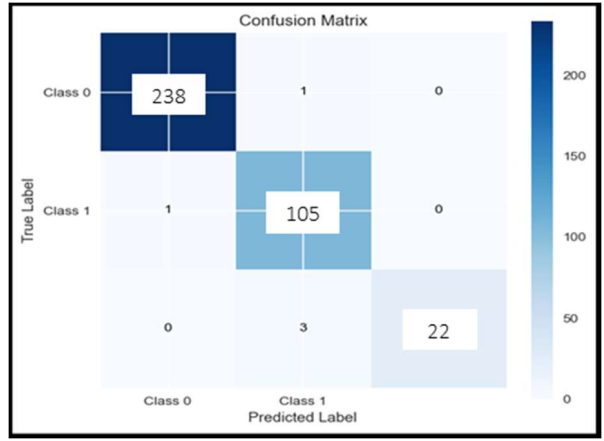

**Figure 3** Confusion Matrix of the XGBoost Model

Figure 3 suggests that Class 0 (Low) shows excellent model performance with 238 True Positives (TP) and 1 instance of False Positives (FP) where Low was misclassified as Medium. Class 1 (Medium) also show Strong performance with 105 TP and 1 FP (Low misclassified as High. Class 2 (High) is the least frequent class, but still has good accuracy (22/25 correct = 88% recall), with 22 TP and 3 FP.

The model performance is also evaluated based on *accuracy*, the *precision*, *recall*, and *F1-score* metrics and the results are shown in Table 4 below.

**Table 4** Classification Report for the XGBoost model Predicting Maintenance Class. .

|  | *Precision* | *Recall* | *F1-score* |
|---:|:---:|:---:|:---:|
| *Low* | 0.98 | 0.98 | 0.99 |
| *Medium* | 0.96 | 0.98 | 0.98 |
| *High* | 0.98 | 0.88 | 0.94 |
| *Accuracy* |  |  | 0.98 |
| *Macro Average* | 0.99 | 0.96 | 0.97 |
| *Weighted Average* | 0.98 | 0.98 | 0.98 |

## 4.3 Model Explainability and Feature Importance Ranking

To enhance transparency and interpretability, SHAP values are computed for the trained classifier to show contribution to the prediction across all samples, based on Shapley values from cooperative game theory [35]. The SHAP value $\emptyset_j$ for feature j is defined as:

$$\emptyset_j = \sum_{S \leq F\{j\}} \frac{|S|!(|F|-|S|-1)!}{|F|!} [f_{S \cup \{j\}}(x) - f_S(x)] \qquad (7)$$

Where F is the set of all features and S represents a subset of features excluding j. The mean absolute SHAP value for each feature is calculated to determine its global influence on model predictions. In infrastructure applications, SHAP has been used to develop reliability indices for power systems [36], flood risk scores in urban planning [37] and drought resilience indices in agricultural management [38]. Figure 4 below shows the global features ranking of the variable importance.

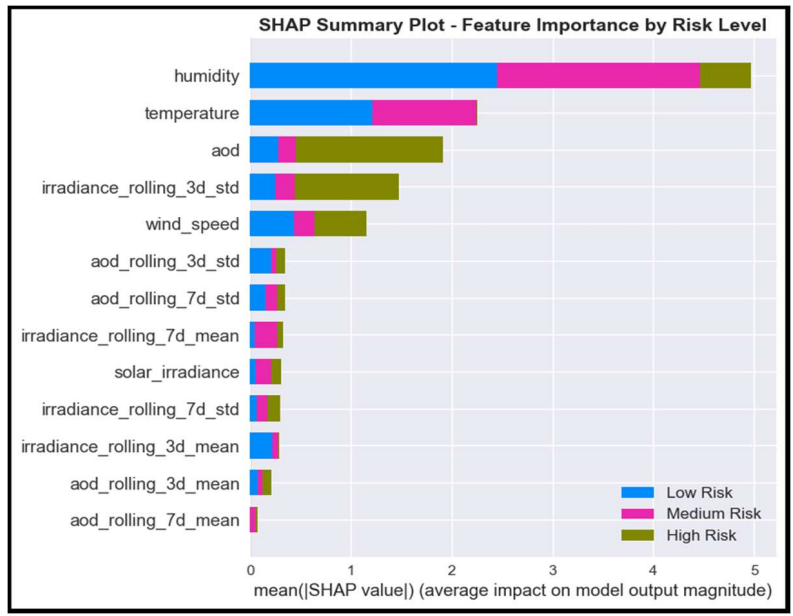

**Figure 4** Global features ranking of the variable importance.

Figure 4 shows that Humidity, temperature and AOD are the dominant predictive factors across all risk levels. Also, Irradiance variability matters more than absolute irradiance as can be seen from the rolling standard deviations ranking higher than absolute irradiance values. This suggests that solar resource consistency is more important for risk prediction than peak solar availability. Wind speed shows moderate importance. Furthermore, SHAP waterfall plots were used to explain individual predictions such as shown in Figure 5 below. These plots decomposed the prediction for a specific sample into its component feature contributions. This provided human-understandable explanations of why a risk prediction is classified as high, medium, or low risk. A very important component for infrastructure planners and regulatory bodies.

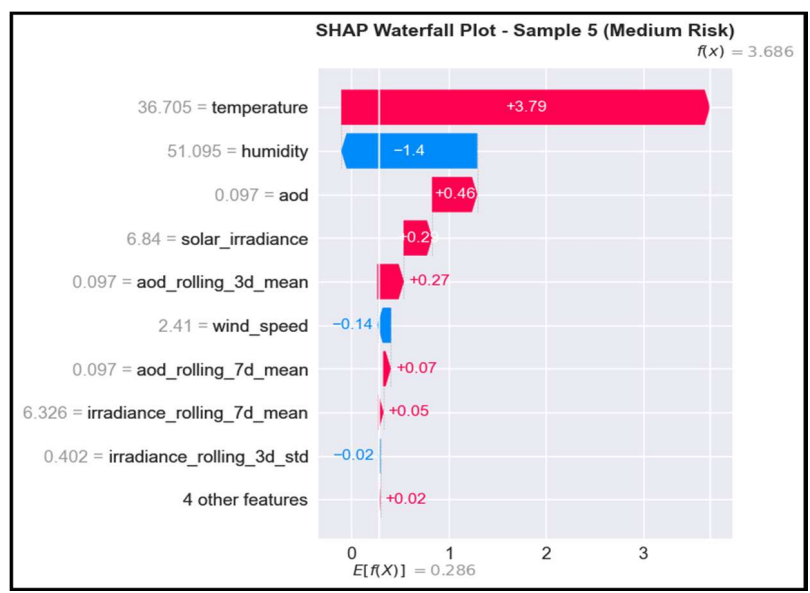

**Figure 5** Features Ranking of Individual prediction

### 4.4 Time Series Forecasting of Maintenance Pressure using Prophet

Figure 6 below displays a two-year (104-week) forecast of the weekly maintenance score, generated using the Prophet model enhanced with lagged and smoothed meteorological regressors. The blue line represents the predicted maintenance score, while the shaded region indicates the uncertainty interval (confidence bounds). The model captures seasonal fluctuations and suggests a gradual upward trend in maintenance risk over the forecast horizon, with periodic spikes reflecting potential high-risk periods. The observed values (black dots) align reasonably well with historical predictions, indicating the model's reliability in learning past dynamics.

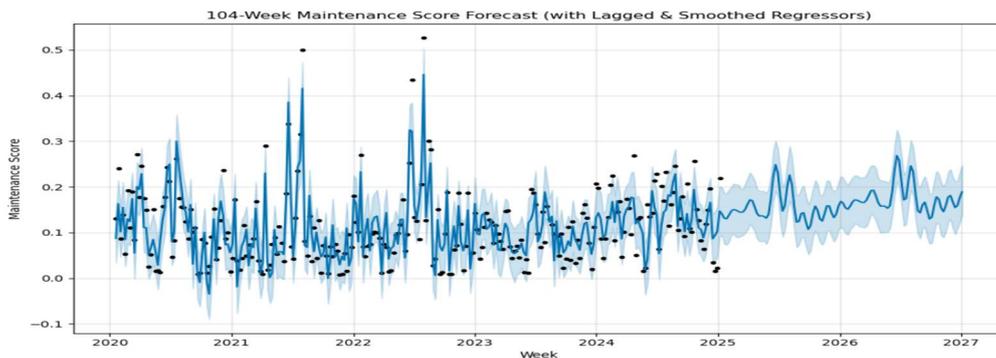

**Figure 6** 2 year (104-week forecast of the weekly maintenance score, using Prophet Model

The Prophet forecasting model achieved a Mean Absolute Error (MAE) of 0.0458, a Root Mean Squared Error (RMSE) of 0.0576, and an $R^2$ score of 0.2498. The low error values indicate strong predictive accuracy, especially considering the limited input space. The $R^2$ indicates that 25% of the changes in maintenance pressure can be explained solely by these environmental factors. This shows the significance of environmental conditions on maintenance risk. It is expected that the inclusion of additional operational or contextual variables will further improve predictive performance.

### 4.5 Green Hydrogen Risk Intelligence Dashboard

To enhance the practical utility of the research findings, the predictive models were prototyped into an interactive risk intelligence dashboard shown in figure 7 below. The dashboard features a Scenario Builder which enable users to adjust time frame and instantly visualise the Maintenance Pressure Risk.

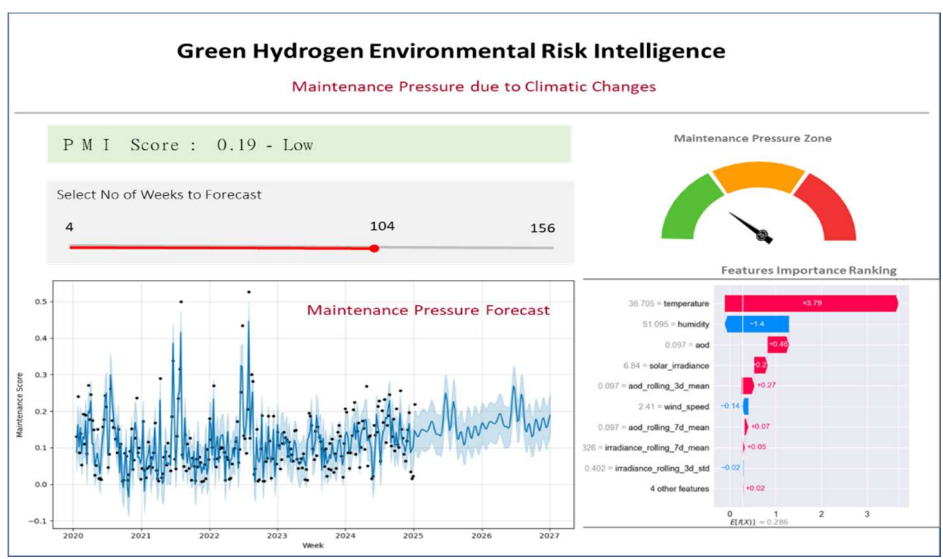

**Figure 7** Green Hydrogen Environmental Risk Intelligence Prototype Dashboard

# 5. Conclusion, Industry and Policy Implications

**Conclusion**

Duqm's 300 km² GH2 auction zone is geographically homogeneous, but meteorological fluctuations across time can impose varying levels of wear and downtime risk on hydrogen production systems. This study developed and validated a composite Maintenance Pressure Index (MPI) as a time-variant, interpretable index for assessing environmental stress and operational risk in green hydrogen infrastructure. By leveraging publicly available NASA POWER and MODIS data for the Duqm hydrogen zone, we developed a rule-based MPI informed by empirical thresholds for aerosol optical depth (AOD), temperature, humidity, wind speed, and irradiance variability. The resulting score was classified into risk levels (Low, Medium, and High) and modelled using XGBoost classifier, with SHAP explainability. The MPI successfully captures real-time environmental volatility and highlights the operational burden posed by weather-driven stressors across the hydrogen production lifecycle.

**Industry Implications**

In the context of Oman's hydrogen industrialization drive and Duqm's 300 km² hydrogen auction zone, this study provides a adaptable, data-driven tool for predictive maintenance planning and risk-readiness assessment. The MPI can be deployed within an investor or operator's dashboard to anticipate adverse environmental conditions and reduce unplanned downtime, especially in systems sensitive to particulate soiling, thermal stress, or wind-induced fatigue. More importantly, the MPI offers a mechanism for infrastructure asset integrity forecasting, which enables early intervention on production facilities. For developers bidding in competitive hydrogen auctions, the MPI provides a quantifiable edge in operational readiness and risk disclosures, thereby enhancing confidence of financial institutions partners. By embedding temporal maintenance scoring into routine operations, plant managers can optimize servicing schedules, extend asset life, and reduce lifecycle costs.

**Policy Implications**

From a regulatory and governance perspective, this study supports a paradigm shift towards dynamic risk-aware regulation. The MPI enables government agencies to monitor evolving environmental stressors and proactively issue "hydrogen weather alerts" or adaptive compliance guidance in periods of elevated maintenance pressure. Furthermore, this opens the door for risk-differentiated auction frameworks, where bidders are scored not only on technical and financial metrics but also on their risk-readiness strategies against MPI trends. The government can integrate MPI risk bands into auction contracts, incentivizing robust engineering and adaptation strategies through bonus mechanisms or flexible delivery milestones. In addition, MPI trends can support policy benchmarking, enabling national planners to track the long-term impact of dust mitigation programs, coastal infrastructure adaptation, or renewable micro grid configurations.

**Future Work**

Future iterations could compute MPI dynamically using a SHAP-weighted additive formulation. Also, consider Multi-Regional Validation to test the framework across diverse geographical and climatic conditions